\documentclass[fleqn,11pt]{article}
\usepackage{jmlr2eNoEditor}

                                                                                                       \usepackage[ruled, oldcommands]{algorithm2e}
\usepackage{stmaryrd}
                                                                                                                                                      \usepackage{amsmath}
                                              \usepackage{amsfonts}
                                              \usepackage{verbatim}
                                              \usepackage{float}
                                              \usepackage{appendix}
                                              \usepackage{listings}
                                              \usepackage{verbatim}
                                              \usepackage[usenames]{color}
                                              \usepackage{graphicx}
                                              \usepackage[
                                              draft
                                              ]{movie15}
                                              \usepackage{ifthen}
                                                \usepackage{alltt}
                                                \usepackage[mathscr]{eucal}
                                                \usepackage{amsfonts}
                                                \usepackage{subfigure}
                                                \usepackage{hhline}
                                                                                                     \usepackage{multirow}
                                                                                                      \usepackage{booktabs}
                                                                                                      \usepackage{colortbl}
                                                                                                      \usepackage{array}
                                                                                                      \usepackage{url}
                                                                                                     \usepackage[colorlinks=true,linkcolor=black,citecolor=black]{hyperref}
                                                                                                      \newfloat{Animation}{p}{ext}
                                                                                                      \usepackage{appendix}

\date{}

                                              \newtheorem{conclusion}{Conclusion}
                                              \newtheorem{lem}{Lemma}
                                              \newtheorem{thm}{Theorem}
                                              \newtheorem{cor}{Corollary}

                                              \newtheorem{clam}{Claim}

                                                                                                                                                                                                                \title{Explaining Adaptation in Genetic Algorithms\\ With Uniform Crossover: The Hyperclimbing Hypothesis}

\author{\name Keki M. Burjorjee\email kekib@cs.brandeis.edu
                                                                                                        }
\editor{}

\begin{document}

\maketitle


\begin{abstract}

The \emph{hyperclimbing hypothesis} is a hypothetical explanation for adaptation in genetic algorithms with uniform crossover (UGAs). Hyperclimbing is an intuitive, general-purpose, non-local search heuristic applicable to discrete product spaces with rugged or stochastic cost functions. The strength of this heuristic lies in its insusceptibility to local optima when the cost function is deterministic, and its tolerance for noise when the cost function is stochastic. Hyperclimbing works by decimating a search space, i.e. by iteratively fixing the values of small numbers of variables. The hyperclimbing hypothesis holds that UGAs work by implementing \emph{efficient} hyperclimbing. Proof of concept for this hypothesis comes from the use of a novel analytic technique involving the exploitation of algorithmic symmetry. We have also obtained experimental results that show that a simple tweak inspired by the hyperclimbing hypothesis dramatically improves the performance of a UGA on large, random instances of MAX-3SAT and the Sherrington Kirkpatrick Spin Glasses problem.
\end{abstract}

\section{Introduction}

Over several decades of use in diverse scientific and engineering fields, evolutionary optimization has acquired a reputation for being a kind of universal acid---a general purpose approach that routinely procures useful solutions to optimization problems with rugged, dynamic, and stochastic cost functions over search spaces consisting of strings, vectors, trees, and instances of other kinds of data structures  \citep{fogel2006evolutionary}. Remarkably, the means by which evolutionary algorithms work is still the subject of much debate. An abiding mystery of the field is the  widely observed utility of genetic algorithms with uniform crossover \citep{Syswerda89,rudnick1994asg,pelikanSherrington,LiMo}. The use of uniform crossover  \citep{ackley1987cmg,Syswerda89} in genetic algorithms causes genetic loci to be unlinked, i.e. recombine freely. It is generally acknowledged that the adaptive capacity of genetic algorithms with this kind of crossover  cannot be explained within the rubric of the \emph{building block hypothesis}, the reigning explanation for adaptation in genetic algorithms with strong linkage between loci \citep{desInnov}. Yet, no alternate, scientifically rigorous explanation for adaptation in genetic algorithms with uniform crossover (UGAs) has been proposed. The \emph{hyperclimbing hypothesis}, presented in this paper, addresses this gap. This hypothesis holds that UGAs perform adaptation by implicitly and efficiently implementing a global search heuristic called \emph{hyperclimbing}.

If the hyperclimbing hypothesis is sound, then the UGA is in good company. Hyperclimbing belongs to a class of heuristics that perform \emph{global decimation}. Global decimation, it turns out, is the state of the art approach to solving large, hard instances of SAT \citep{kroc2009message}.  Conventional global decimation strategies---e.g. Survey Propagation  \citep{mezard2002analytic}, Belief Propagation, Warning Propagation  \citep{journals/corr/cs-CC-0212002}---use message passing algorithms to obtain statistical information about the space being searched. This information is then used to fix the values of one, or a small number, of search space attributes, effectively reducing the size of the search space. The decimation strategy is then recursively applied to the smaller search space. And so on. Survey Propagation, perhaps the best known global decimation strategy, has been used along with \emph{Walksat}  \citep{selman1993local} to solve instances of SAT with upwards of a million variables. The hyperclimbing hypothesis holds that in practice, UGAs also perform adaptation by decimating the search spaces to which they are applied. Unlike conventional decimation strategies, however, a UGA obtains statistical information about the search space \emph{implicitly}, by means other than message passing. 

The rest of this paper is organized as follows: Section \ref{hyperclimbingExplained} provides an informal description of the hyperclimbing heuristic. A more formal description appears in Section A of the online appendix. Section \ref{proofOfConcept}, presents proof of concept, i.e. it describes a stochastic fitness function\footnote{A fitness function is nothing but a cost function with a small twist: the goal is, not to minimize fitness, but to maximize it.} on which a UGA behaves as described in the hyperclimbing hypothesis.  Exploiting certain symmetries inherent within uniform crossover and a containing class of fitness functions, we argue that the adaptive capacity of a UGA scales extraordinarily well as the size of the search space increases. We follow up with experimental tests that validate this conclusion. One way for the hyperclimbing hypothesis to gain credibility is by inspiring modifications to the genetic algorithm that improve performance. Section \ref{validation} presents the results of experiments that show that a simple tweak called \emph{clamping}, inspired by the hyperclimbing hypothesis, dramatically improves the performance of a genetic algorithm on large, randomly generated instances of MAX-3SAT, and the Sherrington Kirkpatric Spin Glasses problem. While not conclusive, this validation does lend considerable support to the hyperclimbing hypothesis\footnote{Then again, \emph{no} scientific theory can be \emph{conclusively} validated. The best one can hope for is \emph{pursuasive} forms of validation \citep{logicofscientificdisc,conjrefut}.}. 
We conclude in Section \ref{conclusion} with a brief discussion of the generalizability of the hyperclimbing hypothesis and its ramifications for Evolutionary Computation and Evolutionary Biology. 

\section{The Hyperclimbing Heuristic} \label{hyperclimbingExplained}

For a sketch of the workings of a hyperclimbing heuristic, consider a search space $S=\{0,1\}^\ell$, and a (possibly stochastic) fitness function that maps points in $S$ to real values. Let us define the \emph{order} of a schema partition \cite{Mitchell:1996:IGA} to simply be the order of the schemata that comprise the partition. Clearly then, schema partitions of lower order are coarser than schema partitions of higher order. The \emph{effect} of a schema partition is defined to be the variance of the expected fitness of the constituent schemata under sampling from the uniform distribution over each schema. So for example, the effect of the schema partition $\#**\#**=\{0**0**,\, 0**1**,\, 1**0**,\, 1**1**\}$ is \[\frac{1}{4}\sum_{i=0}^1\sum_{j=0}^1(F(i**j**)-F(******))^2\] where the operator $F$ gives the expected fitness of a schema under sampling from the uniform distribution. A hyperclimbing heuristic starts by sampling from the uniform distribution over the entire search space. It subsequently identifies a coarse schema partition with a non-zero effect, and limits future sampling to a schema in this partition with above average expected fitness. In other words the hyperclimbing heuristic fixes the defining bits \cite{Mitchell:1996:IGA} of this schema in the population. This schema constitutes a new (smaller) search space to which the hyperclimbing heuristic is recursively applied. Crucially, the act of fixing defining bits in a population has the potential to ``generate" a detectable non-zero effect in a schema partition that previously had a negligible effect. For example, the schema partition $*\#***\#$ can have a negligible effect, while the schema partition $1\#*0*\#$ has a detectable non-zero effect. A more formal description of the hyperclimbing heuristic can be found in Appendix \ref{formaldescription}. 

At each step in its progression, hyperclimbing is sensitive, not to the fitness value of any individual point, but to the sampling means of relatively coarse schemata. This heuristic is, therefore, \emph{natively} able to tackle optimization problems with stochastic cost functions.  Considering the intuitive simplicity of hyperclimbing, this heuristic has almost certainly been toyed with by other researchers in the general field of discrete optimization. In all likelihood it was set aside each time because of the seemingly high cost of implementation for all but the smallest of search spaces or the coarsest of schema partitions. Given a search space comprised by $\ell$ binary variables, there are ${ \ell\choose o}$ schema partitions of order $o$. For any fixed value of $o$,  ${ \ell\choose o} \in \Omega(\ell^o)$ \citep{clr}. The exciting finding presented in this paper is that UGAs can implement hyperclimbing cheaply for large values of $\ell$, and values of $o$ that are small, but greater than one.

\section{Proof of Concept} \label{proofOfConcept}

We introduce a parameterized stochastic fitness function, called a \emph{staircase function}, and provide experimental evidence that a UGA can perform hyperclimbing on a particular parameterization of this function. Then, using symmetry arguments, we conclude that the running time and the number of fitness queries required to achieve equivalent results scale surprisingly well with changes to key parameters. An experimental test validates this conclusion.

\begin{definition} A staircase function descriptor is a 6-tuple $(h,o,\delta,\ell, L,V)$ where $h$, $o$ and $\ell$ are positive integers such that $ho\leq\ell$, $\delta$ is a positive real number, and $L$ and $V$ are matrices with $h$ rows and $o$ columns such that the values of $V$ are binary digits, and the elements of  $L$ are distinct integers in $[\ell]$. \end{definition}

\begin{algorithm}[tb!]
\dontprintsemicolon
\KwIn{$g$ is a chromosome of length $\ell$}\;
$x\leftarrow$ \mbox{some value drawn from the distribution $\mathcal N(0,1)$}\;
\For{$i\leftarrow1$ to h}{
\eIf{$\Xi_{L_{i:}}(g)=V_{i1}\ldots V_{io}$}
   	{$x\leftarrow x+\delta$}
	{$x\leftarrow x-(\delta/(2^o-1))$\;
	\textbf{break}
}
}
\KwRet{$x$}\;

\caption[boo]{\\A staircase function with descriptor $(h,o,\delta,\sigma,\ell,L,V)$\label{staircasealg}}
\end{algorithm}

For any positive integer $\ell$, let $[\ell]$ denote the set $\{1,\ldots,\ell\}$, and let $\mathfrak B_\ell$ denote the set of binary strings of length $\ell$. Given any $k$-tuple, $x$, of integers in $[\ell]$, and any binary string $g\in\mathfrak B_\ell$, let $\Xi_x(g)$ denote the string $b_1, \ldots, b_k$ such that for any $i\in[k]$, $b_i=g_{x_i}$. For any $m\times n$ matrix $M$, and any $i\in[m]$, let $M_{i:}$ denote the $n$-tuple that is the $i^{th}$ row of $M$. Let $\mathcal N(a,b)$ denote the normal distribution with mean $a$ and variance $b$. Then the function, $f$, described by the staircase function descriptor $(h, o, \delta,\ell, L, V)$ is the stochastic function over the set of binary strings of length $\ell$ given by Algorithm \ref{staircasealg}. The parameters $h, o, \delta$, and $\ell$ are called the \emph{height, order, increment} and \emph{span}, respectively, of $f$. For any $i\in[h]$, we define \emph{step} $i$ of $f$ to be the schema $\{g\in\mathfrak B_\ell| \Xi_{L_{i:}}(g)=V_{i1}\ldots V_{io}\}$, and define \emph{stage} $i$ of $f$ to be the schema $\{g\in\mathfrak B_\ell| (\Xi_{L_{1:}}(g)=V_{11}\ldots V_{1o})\wedge\ldots\wedge(\Xi_{L_{i:}}(g)=V_{i1}\ldots V_{io})\}$.
                                                                                                        
A step of the staircase function is said to have been \emph{climbed} when future sampling of the search space is largely limited to that step. Just as it is hard to climb higher steps of a physical staircase without climbing lower steps first, it is computationally expensive to identify higher steps of a staircase function without identifying lower steps first (Theorem 1, Appendix \ref{proofs}). In this regard, it is possible that staircase functions capture a feature that is widespread within the fitness functions resulting from the representational choices of GA users.  The difficulty of climbing step $i\in[h]$ \emph{given} stage $i-1$, however,  is non-increasing with respect to $i$ (Corollary 1, Appendix \ref{proofs}). Readers seeking to ways to visualize staircase functions are refered to Appendix \ref{visualizing}.

                                                  \subsection{UGA Specification} \label{matandmet}
                                                                                                           The pseudocode for the UGA used in this paper is given in Algorithm \ref{sgapseudo}. The free parameters of the UGA are $N$ (the size of the population),  $p_m$ (the per bit mutation probability), and \textsc{Evaluate-Fitness} (the fitness function). Once these parameters are fixed, the UGA is fully specified. The specification of a fitness function implicitly determines the length of the chromosomes, $\ell$. Two points deserve further elaboration:

                                                                                                             \begin{enumerate}
                                                                                                             \item The function \textsc{SUS-Selection} takes a population of size $N$, and a corresponding set of fitness values as inputs. It returns a set of $N$ parents drawn by fitness proportionate \emph{stochastic universal sampling} (SUS). Instead of selecting $N$ parents by spinning a roulette wheel with one pointer $N$ times, stochastic universal sampling selects $N$ parents by spinning a roulette wheel with $N$ equally spaced pointers just once. Selecting parents this way has been shown to reduce sampling error  \citep{Baker:1985:ASM,Mitchell:1996:IGA}.

                                                                                                             \item When selection is fitness proportionate, an increase in the average fitness of the population causes a decrease in selection pressure. The UGA in  Algorithm \ref{sgapseudo} combats this effect by using sigma scaling  \citep[p 167]{Mitchell:1996:IGA} to adjust the fitness values returned by \textsc{Evaluate-Fitness}. These adjusted fitness values, not the raw ones, are used when selecting parents. Let $f^{(t)}_x$ denote the raw fitness of some chromosome $x$ in some generation $t$, and let $\overline{f^{(t)}}$ and  $\sigma^{(t)}$ denote the mean and standard deviation of the raw fitness values in generation $t$ respectively. Then the \emph{adjusted fitness} of $x$ in generation $t$ is given by $h^{(t)}_x$ where, if $\sigma^{(t)}=0$ then $h^{(t)}_x=1$, otherwise,  $$h^{(t)}_x=\min(0,1+\frac{f^{(t)}_x -\overline{f^{(t)}}}{\sigma^{(t)}})$$ 
                                                                                                                 The use of sigma scaling also entails that negative fitness values are handled appropriately.

                                                                                                         \end{enumerate}

.
                                                                                                         \begin{algorithm}[tb!]
                                                                                                                   \dontprintsemicolon
                                                                                                                   \SetLine
                                                                                                                   $pop\leftarrow$ \textsc{Initialize-Population}($N$,$\ell$)\;
                                                                                                                   \While {some termination condition is unreached}{
                                                                                                                   $fitnessValues\leftarrow \textsc{Evaluate-Fitness}(pop)$
                                                                                                                   $adjustedFitVals\leftarrow \textsc{Sigma-Scale}(fitnessValues)$\;
                                                                                                                   $parents\leftarrow\textsc{SUS-Selection}(pop, adjustedFitVals)$\;
                                                                                                                   $crossMasks\leftarrow$\textsc{Generate-UX-Masks}($N/2$, $\ell$)\;
                                                                                                                       \For{i $\leftarrow$ \textrm{1} to $N/2$}{
                                                                                                                            \For{j $\leftarrow$ \textrm{1} to $\ell$}{
                                                                                                                             \eIf{$crossMasks[i,j]=0$}
                                                                                                                                {$newPop[i,j]\leftarrow parents[i,j]$\;
                                                                                                                                  $newPop[i+N/2,j]\leftarrow parents[i+N/2,j]$}
                                                                                                                                {$newPop[i,j]\leftarrow parents[i+N/2,j]$\;
                                                                                                                                  $newPop[i+N/2,j]\leftarrow parents[i,j]$}
                                                                                                                            }
                                                                                                                       }
                                                                                                                       $mutMasks\leftarrow$\textsc{Generate-Mut-Masks($N$, $\ell$, $p_m$)}\;
                                                                                                                       \For{i $\leftarrow$ \textrm{1} to $N$}{
                                                                                                                            \For{j $\leftarrow$ \textrm{1} to $\ell$}{
                                                                                                                                $newPop[i,j]\leftarrow \textsc{xor}(newPop[i,j]$, $mutMasks[i,j])$
                                                                                                                            }
                                                                                                                       }
                                                                                                                       $pop\leftarrow newPop$
                                                                                                                   }

                                                                                                                   \caption[boo]{\label{sgapseudo}Pseudocode for the UGA used.   The population size is an even number, denoted $N$, the length of the chromosomes is $\ell$, and for any chromosomal bit, the probability that the bit will be flipped during mutation (the per bit mutation probability) is $p_m$. The population is represented internally as an $N$ by $\ell$ array of bits, with each row representing a single chromosome.  \textsc{Generate-UX-Masks(}$x,y$) creates an $x$ by $y$ array of bits drawn from the uniform distribution over $\{0,1\}$. \textsc{Generate-Mut-Masks(}$x,y,z$) returns an $x$ by $y$  array of bits such that any given bit is 1 with probability $z$.}
                                                                                                                   \normalsize
                                                                                                             \end{algorithm}

\subsection{Performance\, of \,a UGA on a \,class of\, Staircase Functions} \label{perfof U}

Let $f$ be a staircase function with descriptor $(h,o,\delta,\ell, L, V)$, we say that $f$ is \emph{basic} if $\ell=ho$, $L_{ij}=o(i-1)+j$, (i.e. if $L$ is the matrix of integers from 1 to $ho$ laid out row-wise), and $V$ is a matrix of ones. If $f$ is known to be basic, then the last three elements of the descriptor of $f$ are fully determinable from the first three, and its descriptor can be shortened to $(h,o,\delta)$. Given some staircase function $f$ with descriptor $(h,o,\delta,\ell, L, V)$, we define the \emph{basic form} of $f$ to be the (basic) staircase function with descriptor $(h,o,\delta)$.

\begin{figure*}[tb!]\begin{center}
\subfigure[]{\includegraphics[width=.3\textwidth]{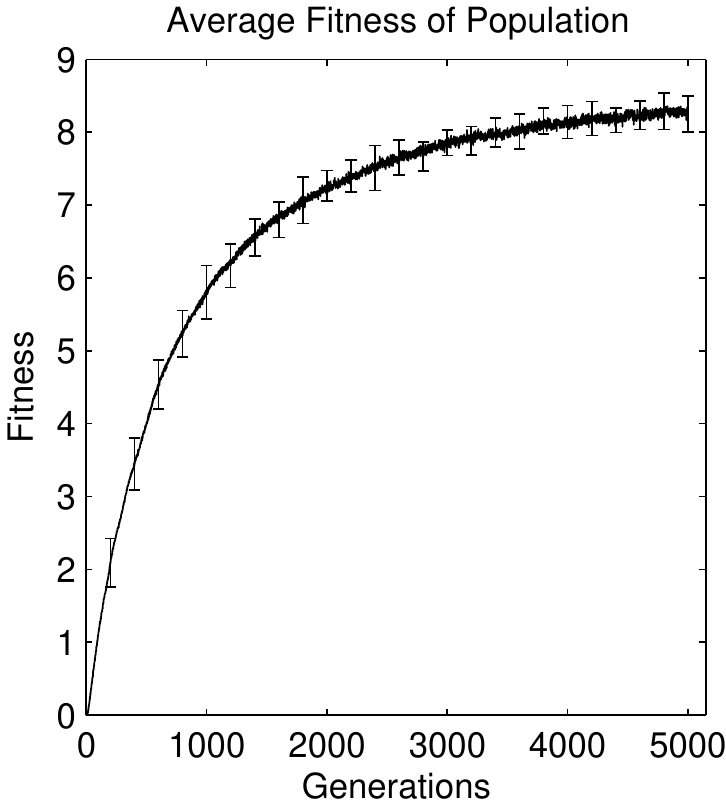}}\quad\quad\quad
\subfigure[]{\includegraphics[width=.3\textwidth]{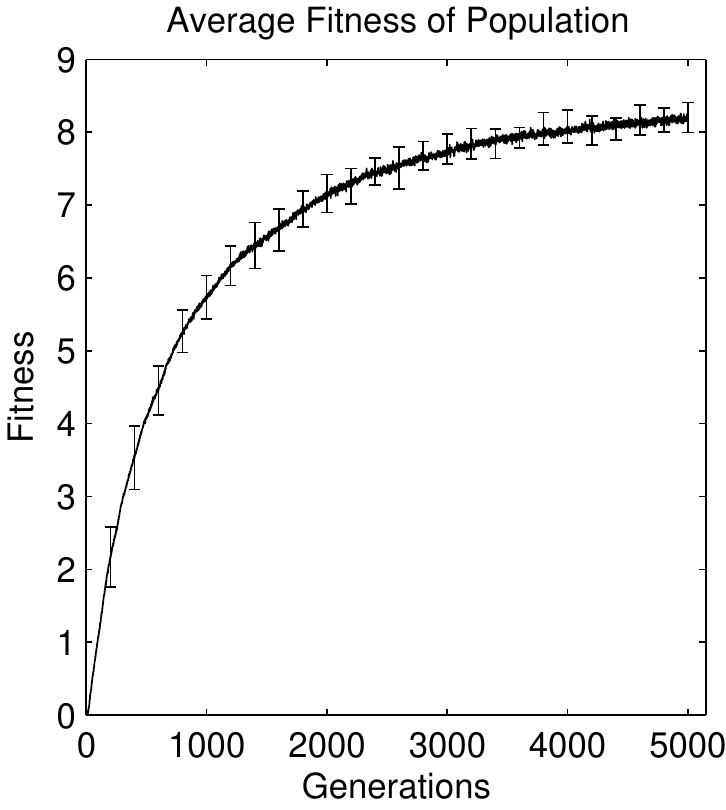}}\end{center}
\begin{center}
\subfigure[]{\includegraphics[width=.30\textwidth]{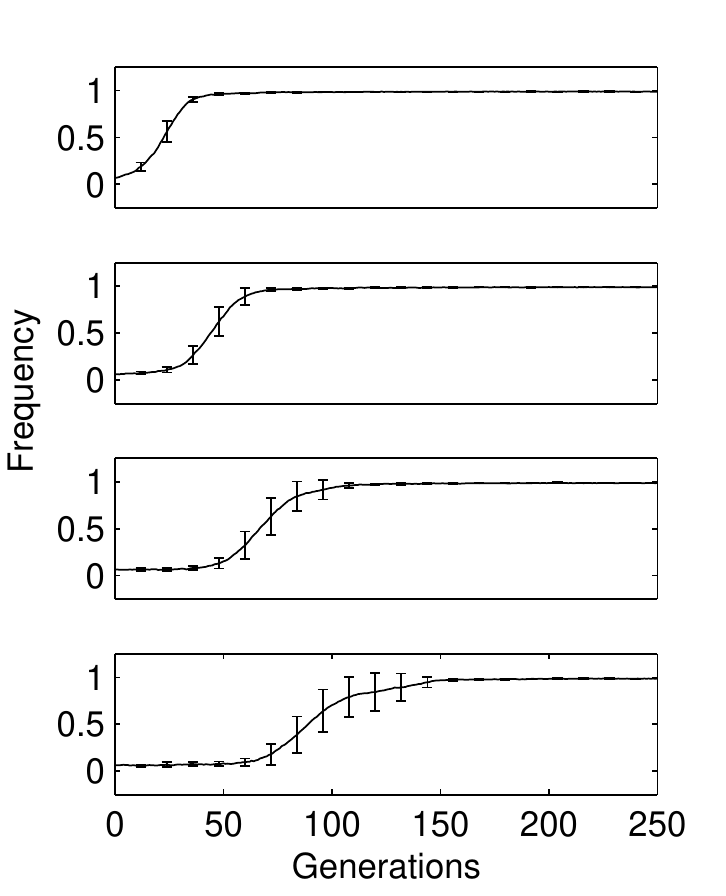}}\quad\quad\quad
\subfigure[]{\includegraphics[width=.30\textwidth]{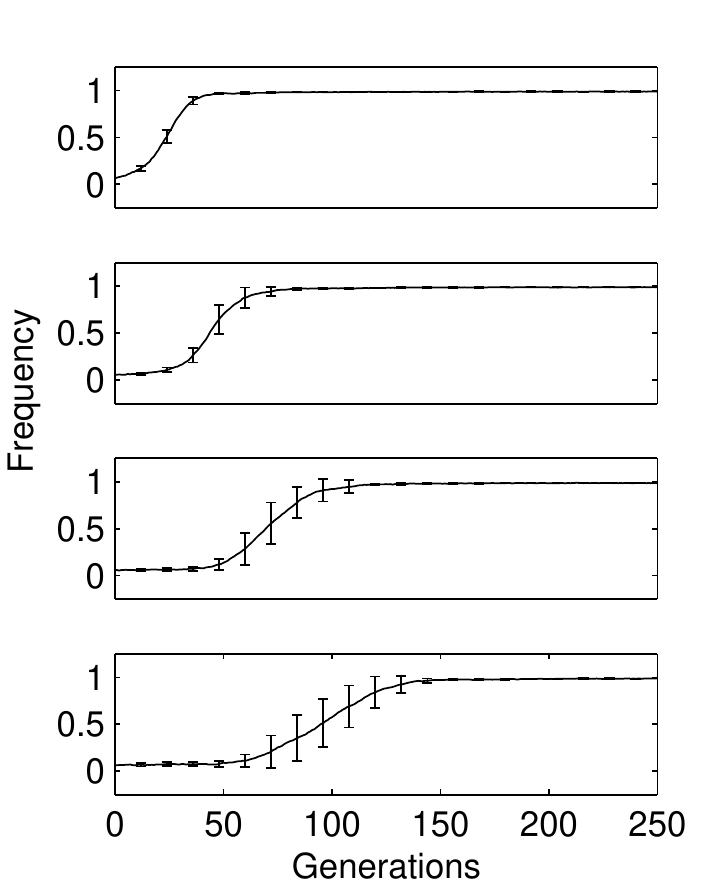}}

\end{center}
\caption{\label{crosstype=0Performance}(a) The mean, across 20 trials, of the average fitness of the population of $U^{\phi^*}i$ in each of 5000 generations.The error bars show five standard errors above and below the mean every $200$ generations. (c) Going from the top plot to the bottom plot, the mean frequencies, across 20 trials, of the first four steps of the staircase function $U^{\phi^*}$  in each of the first 250 generations. The error bars show three standard errors above and below the mean every $12$ generations. (b,d) Same as the plots on the left, but for $U^\phi$}.
\end{figure*}

Let $\phi^*$ be the basic staircase function with descriptor $(h=50, o=4, \delta=0.3)$, and let $U$ denote the UGA defined in section \ref{matandmet} with a population size of  500, and a per bit mutation probability of 0.003 (i.e, $p_m=0.003$).
Figure \ref{crosstype=0Performance}a shows that $U$  is capable of robust adaptation when applied to $\phi^*$ (We denote the resulting algorithm by $U^{\phi^*}$). Figure \ref{crosstype=0Performance}c shows that under the action of $U$, the first four steps of $\phi^*$ go to fixation\footnote{The terms `fixation' and `fixing' are used loosely here. Clearly, as long as the mutation rate is non-zero, no locus can ever be said to go to fixation in the strict sense of the word.}
in ascending order. When a step gets fixed, future sampling will largely be confined to that step---in effect, the hyperplane associated with the step has been climbed. Note that the UGA does not need to ``fully" climb a step before it begins climbing the subsequent step (Figure \ref{crosstype=0Performance}c). 

 \subsection{Symmetry Analysis and Experimental Confirmation}

Formal models of SGAs with finite populations and non-trivial fitness functions \citep{nix1992mga}, are notoriously unwieldy \citep{journals/ec/Holland00}, which is why most theoretical analyses of SGAs assume an infinite population  \citep{DBLP:journals/amai/LiepinsV92, stephens:1999:sebb,conf/gecco/WrightVR03,CoarseGrainingFoga2007}. Unfortunately, since the running time and the number of fitness evaluations required by such models is always infinite, their use precludes the identification of computational efficiencies of the SGA. In the present case, we circumvent the difficulty of formally analyzing finite population SGAs by exploiting some simple symmetries introduced through our definition of staircase functions, and through our use of a crossover operator with no  \emph{positional bias}. The absence of positional bias in uniform crossover  was highlighted by Eshelman et al. \citeyearpar{eshelman1989bcl}. Essentially, permuting the bits of all strings in each generation using some permutation $\pi$ before crossover, and permuting the bits back using $\pi^{-1}$ after crossover has no effect on the dynamics of a UGA. Another way to elucidate this symmetry is by noting that any homologous crossover operator can be modeled as a string of binary random variables. Only in the case of uniform crossover, however, are these random variables all independent and identically distributed.

It is easily seen that loci that are not part of any step of a staircase function are immaterial during fitness evaluation. The absence of positional bias in uniform crossover entails that such loci can also be ignored during recombination. Effectively, then, these loci can be ``spliced out" without affecting the expected average fitness of the population  in any generation. This, and other observations of this type lead to the conclusion below.

Let $W$ be some UGA. For any staircase function $f$, and any $x\in[0,1]$,  let  $p^{(t)}_{(W^f,i)}(x)$ denote the probability that the frequency of \emph{stage} $i$ of $f$ in generation $t$ of $W^{f}$ is $x$. Let $f^*$ be the basic form of $f$. Then, by appreciating the symmetries between the UGAs $W^{f^{*}}$ and $W^{f}$ one can conclude the following:

\begin{conclusion}
For any generation $t$, any $i\in[h]$, and any $x\in[0,1]$, $p^{(t)}_{(W^f,i)}(x)=p^{(t)}_{(W^{f^*},i)}(x)$
\end{conclusion}

This conclusion straightforwardly entails that to raise the average fitness of a population to some attainable value,
\begin{enumerate}
\item The expected number of generations required is \emph{constant} with respect to the span of a staircase function
\item The running time required scales \emph{linearly} with the span of a staircase function
\item The running time and the number of generations are unaffected by the last two elements of the descriptor of a staircase function
\end{enumerate}

Let $f$ be some staircase function with basic form $\phi^*$ (defined in Section \ref{perfof U}). Then, given the above, the application of  $U$ to $f$ should, discounting deviations due to sampling, produce results identical to those shown in Figures \ref{crosstype=0Performance}a and \ref{crosstype=0Performance}c. We validated this ``corollary" by applying $U$ to the staircase function $\phi$ with descriptor $(h=50,o=4, \delta=0.3,\ell=20000, L,V)$ where $L$ and $V$ were randomly generated. The results are shown in Figures \ref{crosstype=0Performance}b and \ref{crosstype=0Performance}d. Note that gross changes to the matrices $L$ and $V$, and an increase in the span of the staircase function by two orders of magnitude did not produce any statistically significant changes. It is hard to think of another algorithm with better scaling properties on this non-trivial class of fitness functions.

\section{Validation}\label{validation}

Let us pause to consider a curious aspect of the behavior of $U^{\phi^*}$. Figure 1 shows that the growth rate of the average fitness of the population of $U^{\phi^*}$ decreases as evolution proceeds, and the average fitness of the population plateaus at a level that falls significantly short of the maximum expected average population fitness of 15. As discussed in the previous section, the difficulty of climbing step $i$ given stage $i-1$ is non-increasing with respect to $i$. So, given that $U$ successfully identifies the first step of $\phi^*$, why does it fail to identify all remaining steps? To understand why, consider some binary string that belongs to the $i^{th}$ stage of $\phi^*$.  Since the mutation rate of $U$ is 0.003, the probability that this binary string will still belong to stage $i$ \emph{after} mutation is $0.997 ^{io}$. This entails that as $i$ increases,  $U^{\phi^*}$ is less able to ``hold" a population within stage $i$. In light of this observation, one can infer that as $i$ increases the sensitivity of $U$ to the conditional fitness signal of step ${i}$  given stage $i-1$ will decrease. This loss in sensitivity explains the decrease in the growth rate of the average fitness of $U^{\phi^*}$. We call the ``wastage" of fitness queries described here \emph{mutational drag}.

To curb mutational drag in UGAs, we conceived of a very simple tweak called \emph{clamping}. This tweak relies on parameters $\texttt{flagFreqThreshold}\in[0.5, 1]$, $\texttt{unflagFreqThreshold}\in[0.5, \texttt{flagFreqThreshold}]$, and the positive integer $\texttt{waitingPeriod}$.  If the \emph{one-frequency} or the \emph{zero-frequency} of some locus (i.e. the frequency of the bit 1 or the frequency of the bit 0, respectively, at that locus) at the beginning of some generation is greater than $\texttt{flagFreqThreshold}$, then the locus is flagged. Once flagged, a locus remains flagged as long as the one-frequency or the zero-frequency of the locus is greater than $\texttt{unflagFreqThreshold}$ at the beginning of each subsequent generation. If a flagged locus in some generation $t$ has remained constantly flagged for the last $\texttt{waitingPeriod}$ generations, then the locus is considered to have passed our fixation test, and is not mutated in generation $t$. This tweak is called clamping because it is expected that in the absence of mutation, a locus that has passed our fixation test will quickly go to strict fixation, i.e. the one-frequency, or the zero-frequency of the locus will get ``clamped" at one for the remainder of the run.

\begin{figure}[tb!]\begin{center}
\subfigure[]{\includegraphics[width=.28\textwidth]{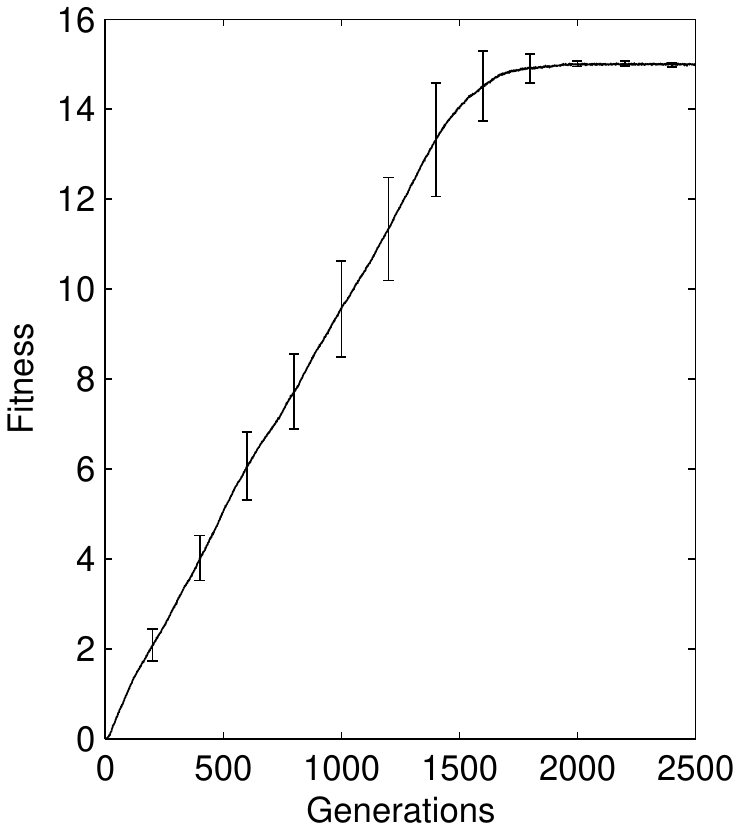}}\quad\quad\quad
\subfigure[]{\includegraphics[width=.28\textwidth]{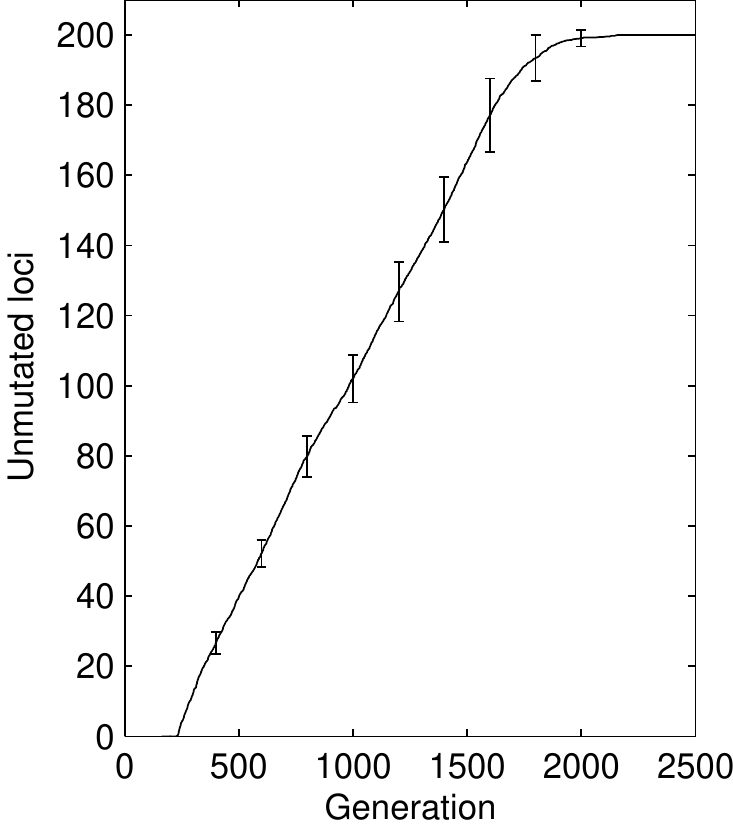}}
\end{center}
\caption{\label{staircaseClamping}\emph{(Top)} The mean (across 20 trials) of the average fitness of the UGA $U_c$ on the staircase function $\phi^*$. Errorbars show five standard errors above and below the mean every 200 generations. \emph{(Bottom)} The mean (across 20 trials) of the number of loci left unmutated by the clamping mechanism. Errorbars show three standard errors above and below the mean every 200 generations}
\end{figure}

Let $U_c$ denote a UGA that uses the clamping mechanism described above and is identical to the UGA $U$ in every other way. The clamping mechanism used by $U_c$ is parameterized as follows: $\texttt{flagFreqThreshold}=0.99$, $\texttt{unflagFreqThreshold}=0.9$, $\texttt{waitingPeriod=200}$. The performance of $U^{\phi^*}_c$ is displayed in figure \ref{staircaseClamping}a. Figure \ref{staircaseClamping}b shows the number of loci that the clamping mechanism left unmutated in each generation. These two figures show that the clamping mechanism effectively allowed $U_c$ to climb all the stages of $\phi^*$. 

If the hyperclimbing hypothesis is accurate, then mutational drag is likely to be an issue when UGAs are applied to other problems, especially large instances that require the use of long chromosomes. In such cases, the use of clamping should improve performance. We now present the results of experiments where the use of clamping clearly improves the performance of a UGA on large instances of MAX-3SAT and the Sherrington Kirkpatrik Spin Glasses problem. 
\subsection{Validation on MAX-3SAT}
MAX-$k$SAT \citep{hoos2004} is one of the most extensively studied combinatorial optimization problems. An instance of this problem consists of $n$ boolean variables, and $m$ clauses. The \emph{literals} of the instance are the $n$ variables and their negations. Each clause is a disjunction of $k$ of the total possible $2n$ literals. Given some MAX-$k$SAT instance, the value of a particular setting of the $n$ variables is simply the number of the $m$ clauses that evaluate to $true$. In a \emph{uniform random} MAX-$k$SAT problem, the clauses are generated by picking each literal at random (with replacement) from amongst the $2n$ literals. Generated clauses containing multiple copies of a variable, and ones containing a variable and its negation, are discarded and replaced.  

Let $Q$ denote the UGA defined in section \ref{matandmet} with a population size of  200 ($N=200$) and a per bit mutation probability of 0.01 (i.e., $p_m=0.01$). We applied $Q$ to a randomly generated instance of the Uniform Random 3SAT problem, denoted $sat$, with 1000 binary variables and 4000 clauses. Variable assignments were straightforwardly encoded, with each bit in a chromosome representing the value of a single variable. The fitness of a chromosome was simply the number of clauses satisfied under the variable assignment represented. Figure \ref{performanceMaxsat}a  shows the average fitness of the population of $Q^{sat}$ over 7000 generations. Note that the growth in the maximum and average fitness of the population tapered off by generation 1000.

The UGA $Q$ was applied to $sat$ once again; this time, however, the clamping mechanism described above was activated in generation 2000. The resulting UGA is denoted $Q^{sat}_c$. The clamping parameters used were as follows: $\texttt{flagFreqThreshold}=0.99$, $\texttt{unflagFreqthreshold}=0.8$, $\texttt{waitingPeriod}=200$. The average fitness of the population of $Q^{sat}_c$ over 7000 generations is shown in Figure \ref{performanceMaxsat}b, and the number of loci that the clamping mechanism left unmutated in each generation is shown in Figure \ref{performanceMaxsat}c. Once again, the growth in the maximum and average fitness of the population tapered off by generation 1000. However, the maximum and average fitness began to grow once again starting at generation 2200. This growth coincides with the commencement of the clamping of loci (compare Figures \ref{performanceMaxsat}b and \ref{performanceMaxsat}c).

\subsection{Validation on an SK Spin Glasses System}
\begin{figure*}[tb!]\begin{center}
	\subfigure[Performance of the UGA $Q^{sat}$]{\includegraphics[width=.3\textwidth]{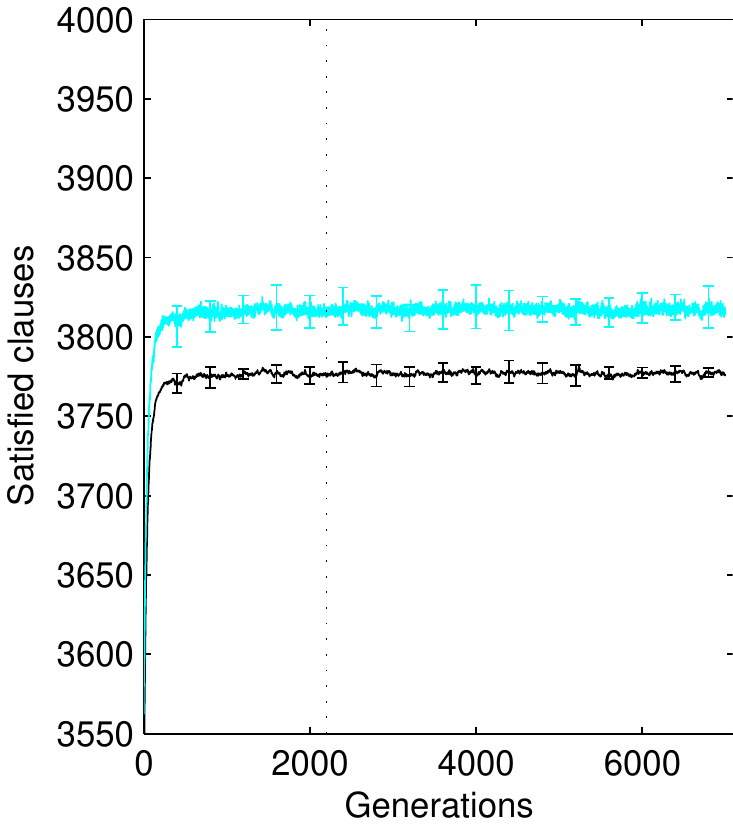}}\quad\quad\quad
\subfigure[Performance of the UGA $Q_c^{sat}$]{\includegraphics[width=.3\textwidth]{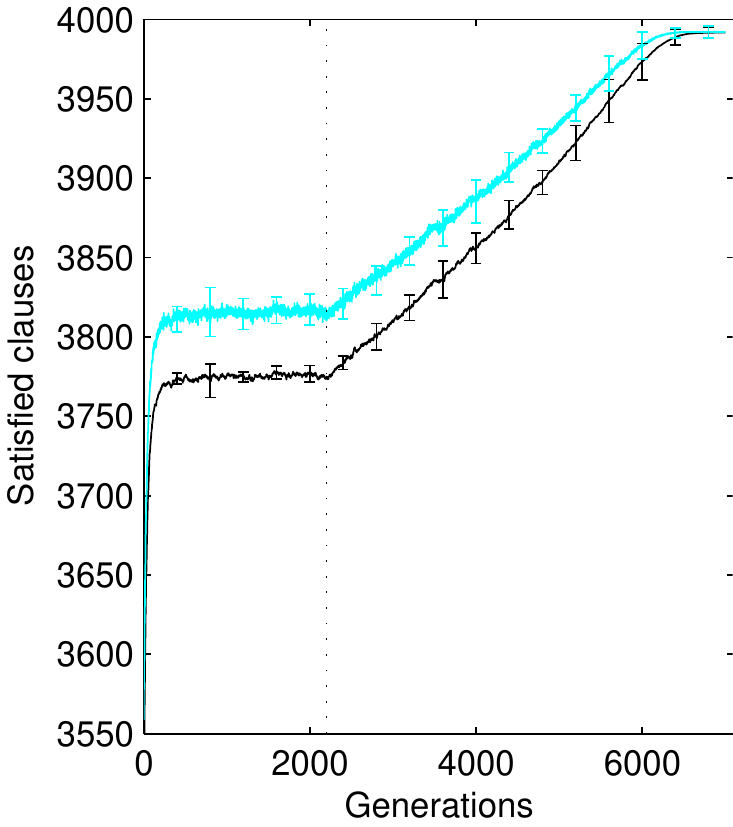}}
\subfigure[Unmutated Loci in UGA $Q_c^{sat}$] {
\includegraphics[width=.3\textwidth]{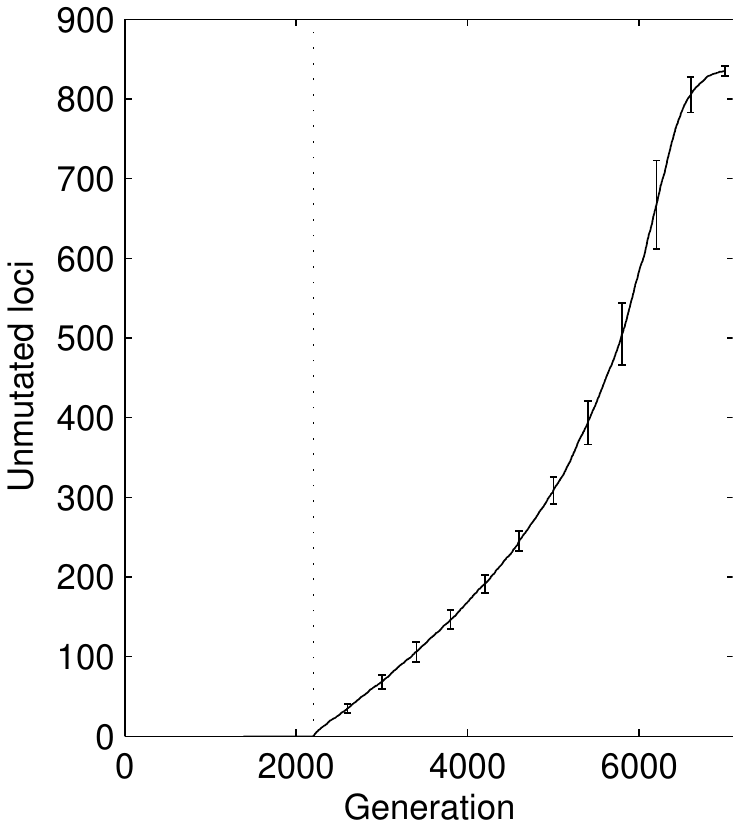}}
\subfigure[Performance of the UGA $Q^{spin}$]{\includegraphics[width=.3\textwidth]{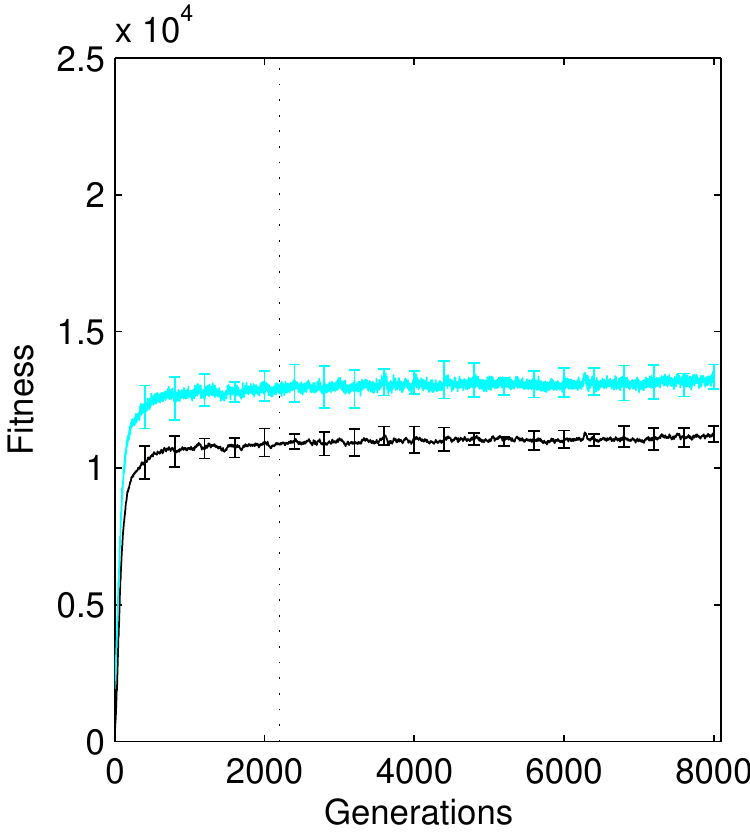}}\quad\quad\quad
\subfigure[Performance of the UGA $Q_c^{spin}$]{\includegraphics[width=.3\textwidth]{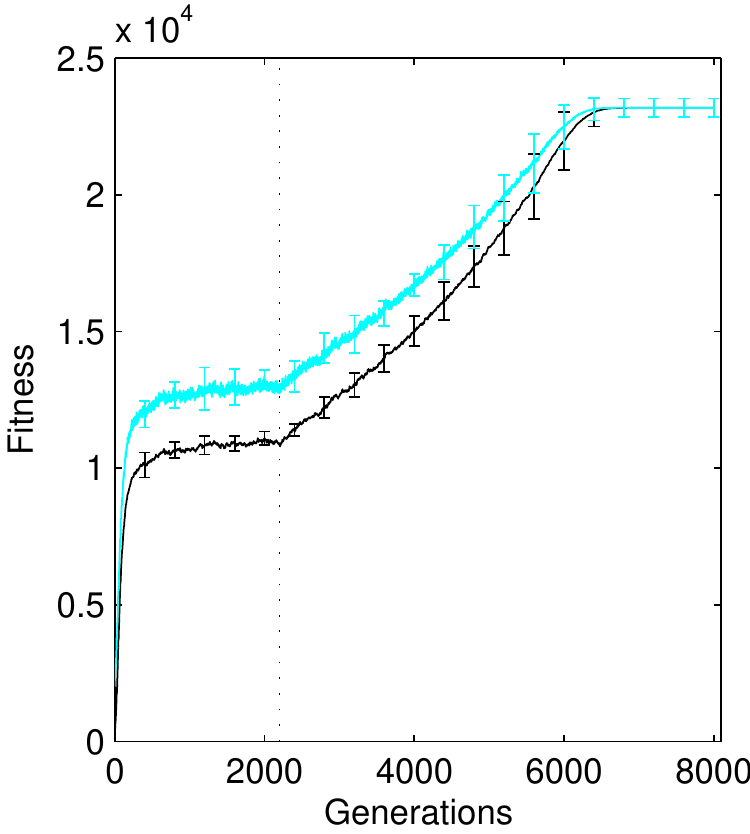}}
\subfigure[Unmutated Loci in UGA $Q_c^{spin}$] {
\includegraphics[width=.3\textwidth]{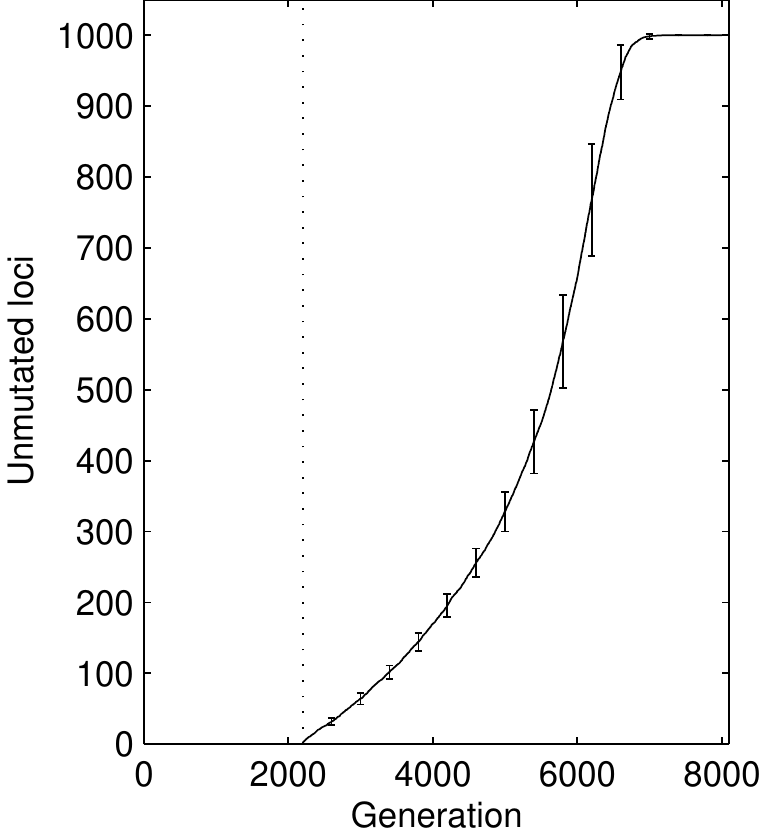}}

\end{center}
\caption{\label{performanceMaxsat}\emph{(a,b)} The performance, over 10 trials, of the UGAs $Q$ and the UGA $Q_c$ on a randomly generated instance of the Uniform Random 3SAT problem with 1000 variables and 4000 clauses. The mean (across trials) of the average fitness of the population is shown in black. The mean of the best-of-population fitness is shown in blue. Errorbars show five standard errors above and below the mean every 400 generations. \emph{(c)} The mean number of loci left unmutated by the clamping mechanism used by $Q_c$. Errorbars show three standard errors above and below the mean every 400 generations. The vertical dotted line marks generation 2200 in all three plots.\emph{(d,e,f)} Same as above, but but for a randomly generated Sherrington Kirkpatrick Spin Glasses System over 1000 spins (see main text for details)}
\end{figure*}

A Sherrington Kirkpatrick Spin Glasses system is a set of coupling constants $J_{ij}$, with $1\leq i < j\leq\ell$. Given a configuration of ``spins" $(\sigma_1,\ldots,\sigma_\ell)$, where each spin is a value in $\{+1,-1\}$, the  ``energy" of the system is given by $$E(\sigma)=-\sum_{1\leq i <j\leq 1}J_{ij}\sigma_i\sigma_j$$. The goal is to find a spin configuration that minimizes energy. By defining the fitness of some spin configuration $\sigma$ to be $-E(\sigma)$ we remain true to the conventional goal in genetic algorithmics of maximizing fitness. The coupling constants in $J$ can either be drawn from the set $\{-1,0,+1\}$, or from the gaussian distribution $\mathcal N(0,1)$. Following Pelikan et al. \citeyearpar{pelikanSherrington}, we used coupling constants drawn from $\mathcal N(0,1)$. Each chromosome in the evolving population straightforwardly represented a spin configuration, with the bits 1 and 0 denoting the spins $+1$ and $-1$ respectively\footnote{Given an $n\times\ell$ matrix $P$ representing a population of $n$ spin configurations, each of size $\ell$, the energies of the spin configurations can be expressed compactly as $-PJ^TP^T$ where $J$ is an $\ell\times\ell$ upper triangular matrix containing the coupling constants of the SK system.}. The UGAs $Q$ and $Q_c$ (described in the previous subsection) were applied to a randomly generated Sherrington Kirkpatrik spin glasses system over 1000 spins, denoted $spin$. The results obtain (Figures \ref{performanceMaxsat}d, \ref{performanceMaxsat}e, and \ref{performanceMaxsat}f) were similar to the results described in the previous subsection.

It should be said that clamping by itself does not cause decimation. It merely enforces strict decimation once a high degree of decimation has already occurred along some dimension. In other words, clamping can be viewed as a decimation ``lock-in" mechanism as opposed to a decimation enforcing mechanism. Thus, the occurrence of clamping shown in Figure \ref{performanceMaxsat} entails the occurrence of decimation. The effectiveness of clamping demonstrated above lends considerable support to the hyperclimbing hypothesis. More support of this kind can be found in the work of Huifang and Mo  \citeyearpar{LiMo} where the use of clamping improved the performance of a UGA on a completely different problem (optimizing the weights of a quantum neural network). A fair portion of the scientific usefulness of these experiments is attributable to the utter simplicity of clamping. Reasoning within the rubric of the hypercling hypothesis, it not difficult to think of adjustments to the UGA that are more effective, but also more complex. From an engineering standpoint the additional complexity would indeed be warranted. From a scientific perspective, however, the additional complexity is a liability because it might introduce suspicion that the adjustments work for reasons other than the one offered here. 

\section{Conclusion} \label{conclusion}
Simple genetic algorithms with uniform crossover (UGAs) perform adaptation by implicitly exploiting one or more features common to the fitness distributions arising in practice. Two key questions are i) What type of features? and ii) How are these features exploited by the UGA (i.e. what heuristic does the UGA implicitly implement)? The hyperclimbing hypothesis is the first scientific theory to venture answers to these questions. In doing so it challenges two commonly held views about the conditions necessary for a genetic algorithm to be effective: First, that the fitness distribution must have a building block structure \citep{desInnov,watsonBook}. Second, that a genetic algorithm will be ineffective unless it makes use of a ``linkage learning" mechanism  \citep{desInnov}. Support for the hyperclimbing hypothesis was presented in the proof of concept and validation sections of this article. Additional support for this hypothesis can be found in i) the weakness of the assumptions undergirding this hypothesis (compared to the building block hypothesis, the hyperclimbing hypothesis rests on weaker assumptions about the distribution of fitness over the search space; see \citealt{myDissertation}), ii) the computational efficiencies of the UGA rigorously identified in an earlier work  
\citep[Chapter 3]{myDissertation}, and iii) the utility of clamping reported by Huifang and Mo \citeyearpar{LiMo}.

If the hyperclimbing heuristic is sound, then the idea of a landscape \citep{WrightLandscape,kauffman1993oos} is not very useful for intuiting the behavior of UGAs. Far more useful is the notion of a \emph{hyperscape}. Landscapes and hyperscapes are both just  ways of conceptualizing fitness functions geometrically. Whereas landscapes draw one's attention to the interplay between the fitness function and the neighborhood structure of individual points, hyperscapes are about the \emph{statistical} fitness properties of individual hyperplanes,  and the ``spatial" relationships between hyperplanes---lower order hyperplanes can \emph{contain} higher order hyperplanes, hyperplanes can \emph{intersect} each other, and disjoint hyperplanes belonging to the same hyperplane partition can be regarded as \emph{parallel}. The use of hyperscapes for intuiting GA dynamics originated with Holland \citeyearpar{holland75:_adapt_natur_artif_system} and was popularized by Goldberg \citeyearpar{Goldberg:1989:GAS}.

Useful as it may be as an explanation for adaptation in UGAs, the ultimate value of the hyperclimbing hypothesis may lie in its \emph{generalizability}. In a previous work 
\citep{myDissertation}, the notion of a unit of inheritance---i.e. a gene---was used to generalize this hypothesis to account for adaptation in simple genetic algorithms with strong linkage between chromosomal loci. It may be possible for the hyperclimbing  hypothesis to be generalized further to account for adaptation in other kinds of evolutionary algorithms,
In general, such algorithms may perform adaptation by efficiently identifying and progressively fixing above average ``aspects"---\emph{units of selection} in evolutionary biology speak---of the chromosomes under evolution. The precise nature of the unit of selection in each case would need to be determined.

If the hyperclimbing hypothesis and its generalizations are sound we would finally have a \emph{unified} explanation for adaptation in evolutionary algorithms. Fundamental advances in the invention, application, and further analysis of these algorithms can be expected to follow. The field of global optimization would be an immediate beneficiary. In turn, a range of fields, including machine learning, drug discovery, and operations research stand to benefit. Take machine learning for instance. Machine learning problems that can be tackled today are, in large part, ones that are reducible \emph{in practice} to convex optimization problems  \citep{journals/jmlr/BennettP06}. The identification of  an intuitive, efficiently implementable, general purpose meta-heuristic for optimization over rugged, dynamic, and stochastic cost functions promises to significantly extend the reach of this field.

Finally, we briefly touch on the interdisciplinary contribution that the hyperclimbing hypothesis makes to a longstanding debate about the units of selection in \emph{\mbox{biological}} populations  \citep{okasha2006eal,extendedPhenotype,selfishGene}. The material presented in the proof of concept section of this paper, and especially the material in Chapter 3 of an earlier work
  \citep{myDissertation} suggest that the most basic unit of selection is, not the individual gene as is commonly thought, but a small \emph{set} of genes. Chapter 3 of the earlier work \citep{myDissertation} 
demonstrates conclusively that as a unit of selection, the latter is not always reducible to instances of the former. In other words, it gives the lie to the common refrain in Population Genetics that multi-gene interactions can be ignored when studying adaptation in biological populations because ``additive effects are the basis for selection"  \citep{wagner02}.
                                                                                                        

\bibliographystyle{plain}

\appendix
\section{The Hyperclimbing Heuristic: Formal Description} \label{formaldescription}
 
 Introducing new terminology and notation where necessary, we present a formal description of the hyperclimbing heuristic. For any positive integer $\ell$, let $[\ell]$ denote the set $\{1,\ldots,\ell\}$, and let  $\mathfrak B_\ell$ denote the set of all binary strings of length $\ell$. For any binary string $g$, let $g_i$ denote the $i^{th}$ bit of $g$. We define the \emph{schema partition model set of $\ell$}, denoted $\mathbb{SPM_\ell}$, to be the power set of $[\ell]$, and define the \emph{schema model set of $\ell$}, denoted $\mathbb{SM_\ell}$, to be the set $\{h:D\rightarrow\{0,1\}|D\in\mathbb{SPM}_\ell\}$. Let $\mathbb S_\ell$ and $\mathbb{SP_\ell}$ be the set of all schemata and schema partitions \cite{Mitchell:1996:IGA}, respectively, of the set $\mathfrak B_\ell$. Given some schema $\gamma\subset\mathfrak B_\ell$, let $\pi(\gamma)$ denote the set $\{i\in[\ell]\,|\, \forall x,y\in\gamma, x_i=y_i\}$.  We define a \emph{schema modeling function} $\textbf{SMF}_\ell:\mathbb S_\ell\rightarrow\mathbb{SM_\ell}$ as follows: for any $\gamma\in\mathbb S_\ell$, $\textbf{SMF}_\ell$ maps $\gamma$ to the function $h:\pi(\gamma)\rightarrow\{0,1\}$ such that for any $g\in \gamma$ and any $i\in\pi(\gamma)$, $h(i)=g_i$.  We define a \emph{schema partition modeling function} $\textbf{SPMF}_\ell:\mathbb{SP_\ell}\rightarrow\mathbb{SPM_\ell}$ as follows: for any $\Gamma\in\mathbb{SP_\ell}$, $\textbf{SPMF}_\ell(\Gamma)=\pi(\gamma)$, where $\gamma\in\Gamma$. As $\pi(\psi)=\pi(\xi)$ for all $\psi,\xi\in\Gamma$, the schema partition modeling function is well defined. It is easily seen that $\textbf{SPF}_\ell$ and $\textbf{SPMF}_\ell$ are both bijective. For any schema model $h\in\mathbb{SM_\ell}$, we denote $\textbf{SMF}_\ell^{-1}(h)$ by $\llbracket h \rrbracket_\ell$. Likewise, for any schema partition model $S\in\mathbb{SPM_\ell}$ we denote $\textbf{SPMF}_\ell^{-1}(S)$ by $\llbracket S\rrbracket_\ell$. Going in the forward direction, for any schema $\gamma\in\mathbb S_\ell$, we denote $\textbf{SMF}_\ell(\gamma)$ by $\langle\gamma\rangle$. Likewise, for any schema partition $\Gamma\in\mathbb{SP_\ell}$, we denote $\textbf{SPMF}_\ell(\Gamma)$ by $\langle\Gamma\rangle$. We drop the $\ell$ when going in this direction, because its value in each case is ascertainable from the operand. For any schema partition $\Gamma$, and any schema $\gamma\in\Gamma$, the \emph{order} of $\Gamma$, and the \emph{order} of $\gamma$ is $|\langle\Gamma\rangle|$.

For any two schema partitions $\Gamma_1, \Gamma_2\in\mathbb{SP_\ell}$, we say that $\Gamma_1$ and $\Gamma_2$ are \emph{orthogonal } if the models of $\Gamma_1$ and $\Gamma_2$ are disjoint (i.e., $\langle \Gamma_1\rangle\cap\langle\Gamma_2\rangle=\emptyset$). Let  $\Gamma_1$ and $\Gamma_2$ be orthogonal schema partitions in $\mathbb{SP_\ell}$, and let $\gamma_1\in\Gamma_1$ and $\gamma_2\in\Gamma_2$ be two schemata. Then the concatenation $\Gamma_1\Gamma_2$ denotes the schema partition  $\llbracket\langle\Gamma_1\rangle \cup\langle\Gamma_2\rangle\rrbracket_\ell$, and the concatenation $\gamma_1\gamma_2$ denotes the schema  $\llbracket h:\langle\Gamma_1\rangle \cup\langle\Gamma_2\rangle\rightarrow\{0,1\}\rrbracket_\ell$ such that for any $i\in \langle\Gamma_1\rangle$,  $h(i)=\langle\gamma_1\rangle(i)$, and for any $i\in \langle\Gamma_2\rangle$, $h(i)=\langle\gamma_2\rangle(i)$. Since $\langle\Gamma_1\rangle$ and $\langle\Gamma_2\rangle$  are disjoint, $\gamma_1\gamma_2$ is well defined. Let $\Gamma_1$ and $\Gamma_2$ be orthogonal schema partitions, and let $\gamma_1\in\Gamma_1$ be some schema. Then $\gamma.\Gamma_2$ denotes the set $\{\gamma\xi\in\Gamma_1\Gamma_2|\xi\in\Gamma_2\}$.

Given some (possibly stochastic) fitness function $f$ over the set $\mathfrak B_\ell$, and some schema $\gamma\in\mathbb S_\ell$, we define the fitness of $\gamma$, denoted $F^{(f)}_{\gamma}$, to be a random variable that gives the fitness value of a binary string drawn from the uniform distribution over $\gamma$. For any schema partition $\Gamma\in\mathbb{SP}_\ell$, we define the \emph{effect} of $\Gamma$,  denoted $\textbf{Effect}\pmb[\Gamma\pmb]$, to be the variance\footnote{We use variance because it is a well known measure of dispersion. Other measures of dispersion may well be substituted here without affecting the discussion} of the expected fitness values of the schemata of $\Gamma$. In other words, \[\textbf{Effect}\pmb[\Gamma\pmb]=2^{-|\langle\Gamma\rangle|}\sum_{\gamma\in\Gamma}\left(\textbf{E}\pmb[F^{(f)}_\gamma\pmb]-2^{-|\langle\Gamma\rangle|}\sum_{\xi\in\Gamma}\textbf{E}\pmb[F^{(f)}_\xi\pmb]\right)^2\]

Let $\Gamma_1,\Gamma_2\in \mathbb{SP_\ell}$ be schema partitions such that $\langle\Gamma_1\rangle\subset\langle\Gamma_2\rangle$. It is easily seen that $\textbf{Effect} \pmb[\Gamma_1\pmb]\leq\textbf{Effect} \pmb[\Gamma_2\pmb] $. With equality if and only if $F^{((f)}_{\gamma_2}=F^{((f)}_{\gamma_1}$ for all schemata $\gamma_1\in\Gamma_1$ and $\gamma_2\in\Gamma_2$  such that $\gamma_2\subset\gamma_1$. This condition is unlikely to arise in practice; therefore, for all practical purposes, the effect of a given schema partition decreases as the partition becomes coarser. The schema partition $\llbracket\,[l]\,\rrbracket_\ell$ has the maximum effect. Let $\Gamma$ and $\Psi$ be two orthogonal schema partitions, and let $\gamma\in\Gamma$ be some schema . We define the conditional effect of $\Psi$ \emph{given} $\gamma$, denoted $\textbf{Effect}\pmb[\Psi|\gamma\pmb]$, as follows:
                                                                                                        \[\textbf{Effect}\pmb[\Psi|\gamma\pmb]=2^{-|\langle\Psi\rangle|}\sum_{\psi\in\Psi}\left(\textbf{E}\pmb[F^{(f)}_{\gamma\psi}\pmb]-2^{-|\langle\Psi\rangle|}\sum_{\xi\in\Psi}\textbf{E}\pmb[F^{(f)}_{\gamma\xi}\pmb]\right)^2\]

A hyperclimbing heuristic works by evaluating the fitness of samples drawn initially from the uniform distribution over the search space. It finds a coarse schema partition $\Gamma$ with a non-zero effect, and limits future sampling to some schema $\gamma$ of this partition whose average sampling fitness is greater then the mean of the average sampling fitness values of the schemata in $\Gamma$.  By limiting future sampling in this way, the heuristic raises the expected fitness of all future samples. The heuristic limits future sampling to some schema by fixing the defining bits  \cite{Mitchell:1996:IGA} of that schema in all future samples. The unfixed loci constitute a new (smaller) search space to which the hyperclimbing heuristic is then recursively applied. Crucially, coarse schema partitions orthogonal to $\Gamma$ that have undetectable \emph{unconditional} effects, may have detectable effects when conditioned by $\gamma$.

\section{Visualizing Staircase Functions} \label{visualizing}
The stages of a staircase function can be visualized as a progression of nested \emph{hyperplanes}\footnote{A hyperplane is a geometrical representation of a schema  \cite[p 53]{Goldberg:1989:GAS}.}, with hyperplanes of higher order and higher expected fitness nested within hyperplanes of lower order and lower expected fitness. By choosing an appropriate scheme for mapping a high-dimensional hypercube onto a two dimensional plot, it becomes possible to visualize this progression of hyperplanes in two dimensions (Appendix \emph{\ref{visualizing}}).

\begin{definition}
A \emph{refractal addressing system} is a tuple $(m,n, X,Y)$, where $m$ and $n$ are positive integers, and $X$ and $Y$ are matrices with $m$ rows and $n$ columns such that the elements in $X$ and $Y$ are distinct positive integers from the set $[2mn]$, such that for any $k\in[2mn]$, $k$ is in $X \Longleftrightarrow k$ is not in $Y$ (i.e. the elements of $[2mn]$ are evenly split between $X$ and $Y$).
\end{definition}

The refractal addressing system $(m,o,X,Y)$  determines how the set $\mathfrak B_{2mn}$ gets mapped onto a $2^{mn}\times 2^{mn}$ grid of pixels. For any bitstring $g\in\mathfrak B_{2mn}$ the $xy$-address (a tuple of two values, each between 1 and $2^{mn}$) of the pixel representing $g$ is given by Algorithm \ref{fractaladdressing}.

\begin{figure*}[tb!]\begin{center}
\subfigure[]{\includegraphics[width=.45\textwidth]{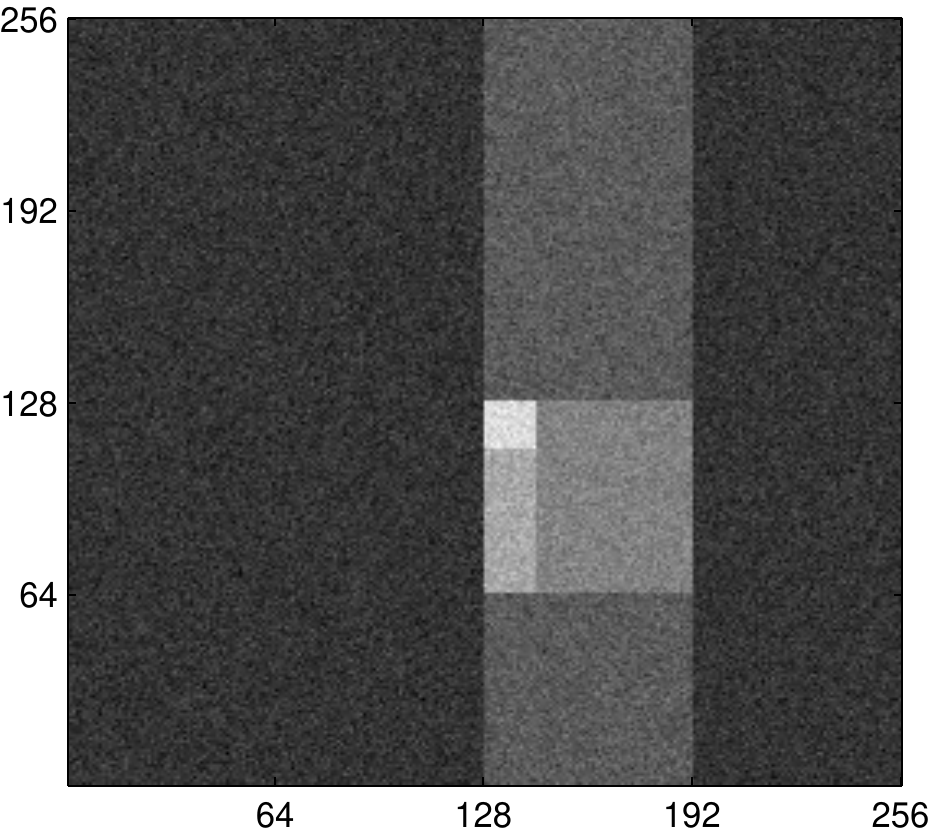}}\quad\quad\quad
\subfigure[]{\includegraphics[width=.45\textwidth]{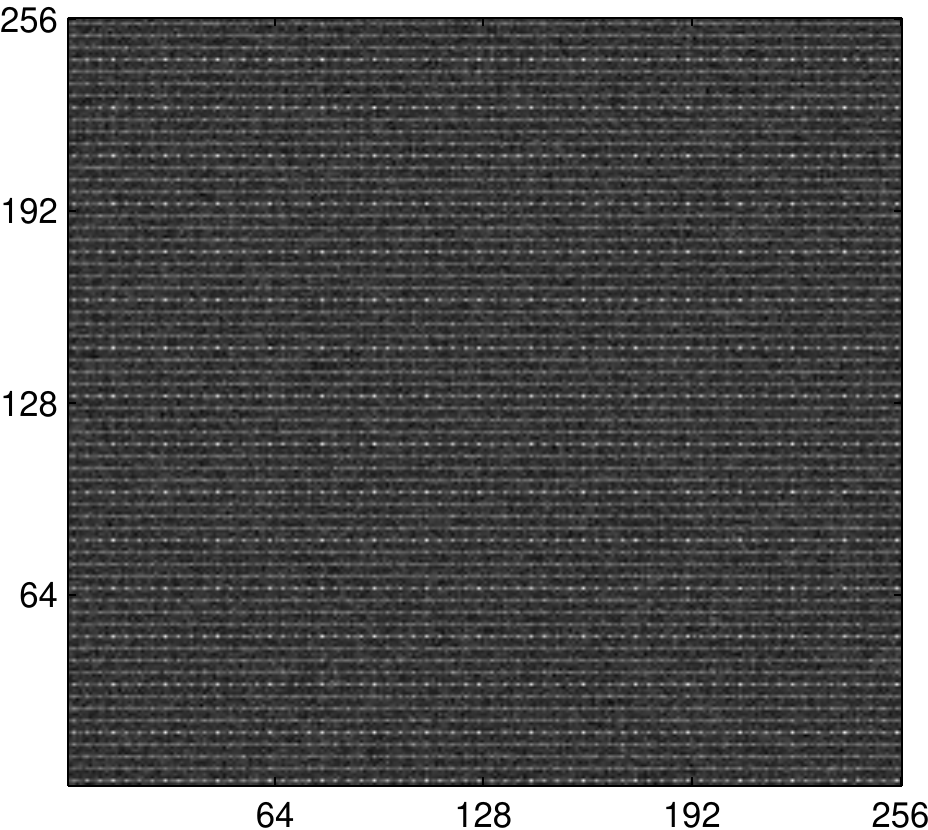}}\end{center}\caption{\label{ladderfunctionvis} A refractal plot of the staircase function $f$ under the refractal addressing systems $A$ (\emph{left}) and $A'$ (\emph{right}).}
\end{figure*}

\noindent\textbf{Example: } Let $(h=4,o=2,\delta=3,\ell=16,L,V )$ be the descriptor of a staircase function $f$, such that \[V=\left[\begin{array}{cc}1&0\\0&1\\0&0\\1&1\end{array}\right]\] Let $A=(m=4,n=2,X,Y)$ be a refractal addressing system such that  $X_{1:}=L_{1:}$,  $Y_{1:}=L_{2:}$, $X_{2:}=L_{3:}$, and $Y_{2:}=L_{4:}$. A \emph{refractal plot}\footnote{The term ``refractal plot" describes the images that result when \emph{dimensional stacking} is combined with \emph{pixelation} \cite{bb12385}.} of $f$ is shown in Figure \ref{ladderfunctionvis}a.

This image was generated by querying $f$ with every bitstring in $\mathfrak B_{16}$, and plotting the resulting fitness value of each chromosome as a greyscale pixel at the chromosome's refractal address under the addressing system $A$. The fitness values returned by $f$ have been scaled to use the full range of possible greyscale shades\footnote{We used the Matlab function \texttt{imagesc()}}. Lighter shades signify greater fitness. The four stages of $f$ can easily be discerned.

Suppose we generate another refractal plot of $f$ using the same addressing system $A$, but a different random number generator seed; because $f$ is stochastic, the greyscale value of any pixel in the resulting plot will then most likely differ from that of its homolog in the plot shown in Figure \ref{ladderfunctionvis}a. Nevertheless, our ability to discern the stages of $f$ would not be affected. In the same vein, note that when specifying $A$, we have not specified the values of the last two rows of $X$ and $Y$; given the definition of $f$ it is easily seen that these values are immaterial to the discernment of its ``staircase structure".
                                                                                                                                                                                                                \begin{algorithm}[tb!]
\dontprintsemicolon
\KwIn{$g$ is a chromosome of length $2mn$}\;
$granularity\leftarrow2^{mn}/2^n$\;
$x\leftarrow1$\;
$y\leftarrow1$\;
\For{$i\leftarrow1$ to $m$}{
	$x\leftarrow x+granularity*\textsc{Bin-To-Int }(\Xi_{X_{i:}}(g))$\;
	$y\leftarrow y+granularity*\textsc{Bin-To-Int }(\Xi_{Y_{i:}}(g))$\;
	$granularity\leftarrow granularity/2^n$\;
}

\KwRet{$x, y$}\;

\caption{\label{fractaladdressing}The algorithm for determining the ($x$, $y$)-address of a chromosome under the refractal addressing system $(m,n,X,Y)$. The function \textsc{Bin-To-Int} returns the integer value of a binary string.}
\end{algorithm}

On the other hand, the values of the first two rows of $X$ and $Y$ are highly relevant to the discernment of this structure.  Figure \ref{ladderfunctionvis}b shows a refractal plot of $f$ that was obtained using a refractal addressing system  $A'=(m=4,n=2, X',Y')$ such that $X'_{4:}=L_{1:}$,  $Y'_{4:}=L_{2:}$, $X'_{3:}=L_{3:}$, and $Y'_{3:}=L_{4:}$. Nothing remotely resembling a staircase is visible in this plot.

The lesson here is that the discernment of the fitness staircase inherent within a staircase function depends critically on how one `looks' at this function. In determining the `right' way to look at $f$ we have used information about the descriptor of $f$, specifically the values of $h,o$, and  $L$. This information will not be available to an algorithm which only has query access to $f$.

\begin{figure*}[tb!]
\begin{center}\subfigure{\includegraphics[width=.45\textwidth]{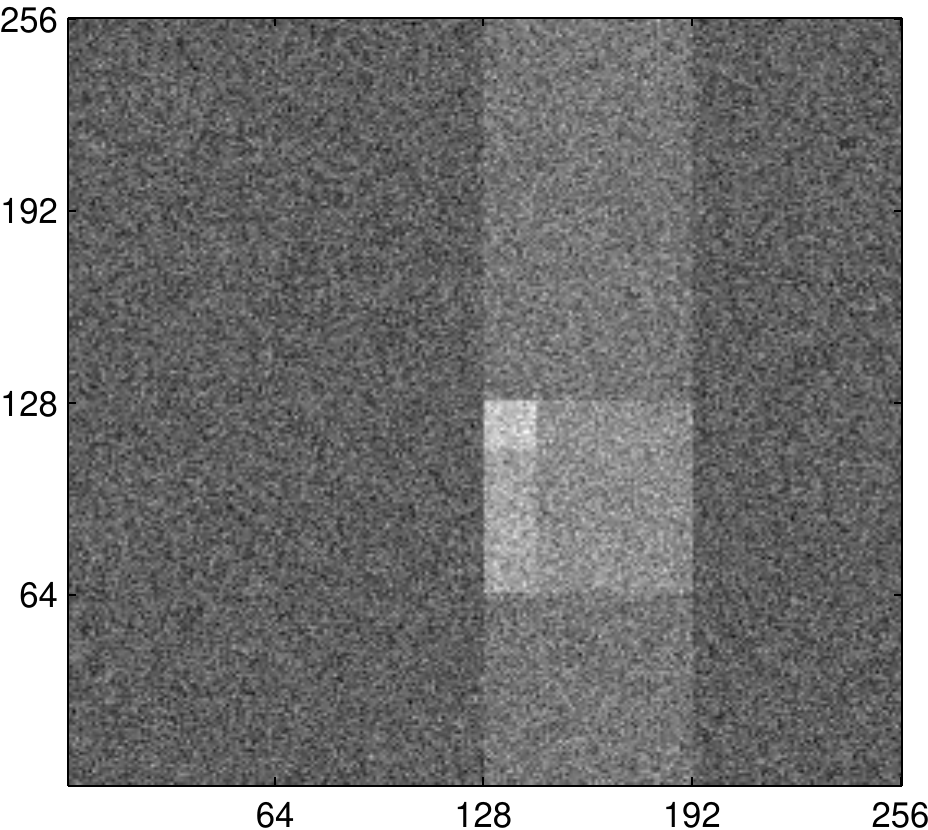}}
\quad\quad\quad\subfigure{\includegraphics[width=.45\textwidth]{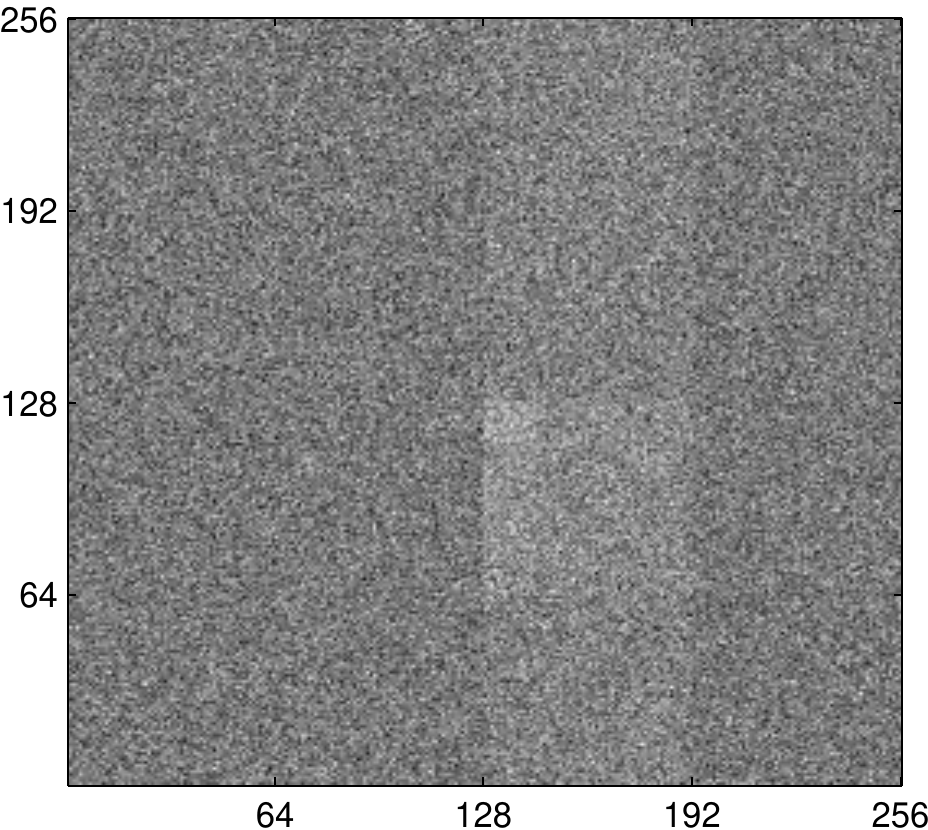}}\end{center}
\caption{\label{fractalPlot2} Refractal plots under $A$ of two staircase functions, which differ from  $f$ only in their increments---1 \emph{(left plot)} and 0.3 \emph{(right plot)}  as opposed to  3.}
\end{figure*}

Even if one knows the right way to look at a staircase function, the discernment of the fitness staircase inherent within this function can still be made difficult by a low value of the increment parameter.  Figure \ref{fractalPlot2} lets us visualize the decrease in the salience of the fitness staircase of $f$ that accompanies a decrease in the increment parameter of this staircase function. In general, a decrease in the increment results in a decrease in the `contrast' between the stages of that function, and an increase the amount of computation required to discern these stages.

\section{Analysis of Staircase Functions}\label{proofs}

Let $\ell$ be some positive integer. Given some (possibly stochastic) fitness function $f$ over the set $\mathfrak B_\ell$, and some schema $\gamma\subseteq\mathfrak{B_\ell}$ we define the \emph{fitness signal} of $\gamma$, denoted $S(\gamma)$,  to be $\textbf{E}\pmb[F^{(f)}_\gamma\pmb]-\textbf{E}\pmb[F^{(f)}_{\mathfrak{B_\ell}}\pmb]$. Let $\gamma_1\subseteq\mathfrak{B_\ell}$ and $\gamma_2\subseteq\mathfrak{B_\ell}$ be schemata in two orthogonal schema partitions.  We define the \emph{conditional fitness signal of}  $\gamma_1$ \emph{given} $\gamma_2$, denoted $S(\gamma_1\,|\,\gamma_2)$, to be the difference between the fitness signal of  $\gamma_1\gamma_2$ and the fitness signal of $\gamma_2$, i.e. $S(\gamma_1\,|\,\gamma_2)=S(\gamma_1\gamma_2)-S(\gamma_2)$. Given some staircase function $f$ we denote the $i^{th}$ step of $f$ by $\lfloor f \rfloor_i$ and denote the $i^{th}$ stage of $f$ by $\lceil f \rceil_i$.

                                                                                                         Let $f$ be a staircase function with descriptor $(h,o,\delta,\ell, L,V)$. For any integer $i\in[h]$, the fitness signal of $\lfloor f \rfloor_i$ is one measure of the difficulty of  ``directly" identifying step  $i$ (i.e., the difficulty of determining step $i$ without first determining any of the preceding steps $1,\ldots,i-1$). Likewise, for any integers $i, j$ in $[h]$ such that $i>j$, the conditional fitness signal of step $i$ given stage $j$ is one  measure of the difficulty of ``directly" identifying step $i$ given stage $j$ (i.e. the difficulty of determining $\lfloor f \rfloor_i$ given $\lceil f \rceil_j$ without first determining any of the intermediate steps $\lfloor f \rfloor_{j+1}, \ldots, \lfloor f \rfloor_{i-1}$.

                                                                                                         For any $i\in[h]$, by Theorem 1 (see below), the unconditional fitness signal of step $i$ is \[\frac{\delta}{2^{o(i-1)}}\] This value decreases exponentially with $i$ and $o$. It is reasonable, therefore, to suspect that the direct identification of step $i$ of $f$  quickly becomes infeasible with increases in  $i$ and $o$. Consider, however, that by Corollary 1, for any $i\in\{2,\ldots,h\}$, the  \emph{conditional} fitness signal of step $i$ \emph{given} stage $(i-1)$ is  $\delta$, a \emph{constant} with respect to $i$. Therefore, if some algorithm can identify the first step of $f$, one should be able to use it to indirectly identify all remaining steps in time and fitness queries that scale linearly with the height of $f$.

                                                                                                       \begin{lem}For any staircase function $f$ with descriptor $(h,o,\delta,\ell, L,V)$, and any integer $i\in[h]$, the fitness signal of stage $i$ is $i\delta $.
                                                                                                       \end{lem}
                                                                                                       \noindent \textsc{Proof:} Let $x$ be the expected fitness of $\mathfrak B_\ell$ under uniform sampling.  We first prove the following claim:
                                                                                                       \begin{clam} \label{fgh}The fitness signal of stage $i$ is $i\delta -x$
                                                                                                       \end{clam}
                                                                                                       The proof of the claim follows by induction on  $i$. The base case, when $i=h$ is easily seen to be true from the definition of a staircase function. For any $k\in\{2,\ldots,h\}$, we assume that the hypothesis holds for $i=k$, and prove that it holds for $i=k-1$.  For any $j\in[h]$, let $\Gamma_j\in\mathbb{SP_\ell}$ denote the schema partition containing step $i$. The fitness signal of stage $k-1$ is given by \[\frac{1}{2^o}\left(S(\lceil f \rceil_k)+\sum_{\psi\,\in\,\Gamma_k\backslash\{\lfloor f\rfloor_k\}}S(\lceil f \rceil_{k-1}\psi)\right)\]
                                                                                                       \[=\frac{k\delta -x}{2^o}+\frac{2^o-1}{2^o}\left(\delta(k-1)-\frac{\delta}{2^o-1}-x\right)\]
                                                                                                       where the first term of the right hand side of the above expression follows from the inductive hypothesis, and the second term follows from the definition of a staircase function. Manipulation of this expression yields \[\frac{k\delta +(2^o-1)\delta(k-1)-\delta -2^ox}{2^o}\] which, upon further manipulation, yields $(k-1)\delta -x$.

                                                                                                       This completes the proof of the claim. To prove the lemma, we must prove that $x$ is zero. By claim \ref{fgh}, the fitness signal of the first stage is $\delta-x$. By the definition of a staircase function then,\[x=\frac{\delta-x}{2^o}+\frac{2^o-1}{2^o}\left(-\frac{\delta}{2^o-1}\right)\]
                                                                                                       Which reduces to
                                                                                                       \[x=-\frac{x}{2^o}\]
                                                                                                       Clearly, $x$ is zero. \quad $\Box$

                                                                                                       \begin{cor} For any $i\in\{2,\ldots,h\}$, the conditional fitness signal of step $i$ given stage $i-1$ is $\delta$
                                                                                                       \end{cor}
                                                                                                       \textsc{Proof} The conditional fitness signal of step $i$ given stage $i-1$ is given by
                                                                                                       \begin{align*}
                                                                                                       &S(\lfloor f\rfloor_i\,\,|\,\,\lceil f \rceil_{i-1})\\
                                                                                                       &=S(\lceil f\rceil_i)-S(\lceil f\rceil_{i-1})\\
                                                                                                       &=(i\delta -(i-1)\delta)\\
                                                                                                       &=\delta\,\,\Box \end{align*}
                                                                                                       \begin{thm} For any staircase function $f$ with descriptor $(h,o,\delta,\sigma,\ell, L,V)$, and any integer $i\in[h]$, the fitness signal of step $i$ is $\delta/2^{o(i-1)}$.
                                                                                                       \end{thm}
                                                                                                       \noindent \textsc{Proof:} For any $j\in[h]$, let $\Lambda_j\in\mathbb{SP_\ell}$ denote of the partition containing stage $j$, and let $\Gamma_j\in\mathbb{SP_\ell}$ denote of the partition containing step $j$.
                                                                                                       We first prove the following claim
                                                                                                       \begin{clam} \label{sfgs}For any $i\in[h]$,  \[\sum_{\xi\,\in\, \Lambda i\backslash\{\lceil f \rceil_i\}}S(\xi)=-i\delta\]\end{clam}
                                                                                                       The proof of the claim follows by induction on $i$. The proof for the base case $(i=1)$ is as follows:
                                                                                                       \[\sum_{\xi\,\in\,\Lambda_1\backslash\{\lceil f\rceil_1\}}S(\xi)=(2^o-1)\left(\frac{-\delta}{2^o-1}\right)=-\delta\]
                                                                                                       For any $k\in[h-1]$ we assume that the hypothesis holds for $i=k$, and prove that it holds for $i=k+1$.
                                                                                                       \[\sum_{\xi\,\in\,\Lambda_{k+1}\backslash\{\lceil f\rceil_{k+1}\}}S(\xi)\]
                                                                                                       \[\phantom{aaaa}=\sum_{\psi\in\Gamma_{k+1}\backslash\{\lfloor f \rfloor_{k+1}\}}S(\lceil f\rceil_k\psi)+
                                                                                                       \sum_{\xi\,\in\,\Lambda_k\backslash\{\lceil f \rceil_k\}}\,\,\sum_{\psi\,\in\,\Gamma_{k+1}}S(\xi\psi)
                                                                                                       \]
                                                                                                       \[\phantom{aaaa}=\sum_{\psi\,\in\,\Gamma_{k+1}\backslash\{\lceil f \rceil_{k+1}\}}S(\lceil f \rceil_k\psi)+
                                                                                                       \sum_{\psi\,\in\,\Gamma_{k+1}}\,\,\sum_{\xi\,\in\,\Lambda_k\backslash\{\lceil f \rceil_k\}}S(\xi\psi)
                                                                                                       \]
                                                                                                       \[\phantom{aaaa}=(2^o-1)S(\lceil f \rceil_k)+
                                                                                                       2^o\left(\sum_{\xi\,\in\,\Lambda_k\backslash\{\lceil f \rceil_k\}}S(\xi)\right)
                                                                                                       \]
                                                                                                       where the first and last equalities follow from the definition of a staircase function. Using Lemma 1 and the inductive hypothesis, the right hand side of this expression can be seen to equal \[(2^o-1)\left(k\delta -\frac{\delta}{2^o-1}\right)-2^o k \delta \]
                                                                                                       which, upon manipulation, yields $-\delta(k+1)$.

                                                                                                       For a proof of the theorem, observe that step 1 and stage 1 are the same schema. So, by Lemma 1,
                                                                                                       $S(\lfloor f \rfloor_1)=\delta$. Thus, the theorem holds for $i=1$. For any $i\in\{2,\ldots,h\}$,
                                                                                                       \[S(\lfloor f \rfloor_i) =\frac{1}{(2^o)^{i-1}}\left(S(\lceil f \rceil_{i})+\sum_{\xi\,\in\,\Lambda_{i-1}\backslash\{\lceil f \rceil_{i-1}\}}S(\xi\lfloor f \rfloor_k)\right)\]
                                                                                                       \[\phantom{S(\lfloor f \rfloor_i)}=\frac{1}{(2^o)^{i-1}}\left(S(\lceil f \rceil_i)+\sum_{\xi\,\in\,\Lambda_{i-1}\backslash\{\lceil f \rceil_{i-1}\}}S(\xi)\right)\]
                                                                                                       where the last equality follows from the definition of a staircase function. Using Lemma 1 and Claim \ref{sfgs}, the right hand side of this equality can be seen to equal
                                                                                                       \[\frac{i\delta  -(i-1)\delta}{(2^o)^{i-1}}\]
                                                                                                       \[=\frac{\delta}{2^{o(i-1)}}\quad\Box\\\]

                                                                                                         \end{document}